\documentclass[final]{cvpr}

\usepackage{times}
\usepackage{epsfig}
\usepackage{graphicx}
\usepackage{amsmath}
\usepackage{amssymb}

\makeatletter
\@namedef{ver@everyshi.sty}{}
\makeatother
\usepackage{tikz}
\usepackage{pgfplots}

\usepackage{color}
\usepackage{xspace}
\usepackage[export]{adjustbox}
\usepackage{siunitx}
\usepackage{subfig}
\usepackage{textcomp}
\usepackage{physics}
\usepackage{mathdots}
\usepackage{yhmath}
\usepackage{cancel}
\usepackage{siunitx}
\usepackage{array}
\newcolumntype{H}{>{\setbox0=\hbox\bgroup}c<{\egroup}@{}}
\usepackage{multirow}
\usepackage{gensymb}
\usepackage{tabularx}
\usepackage{booktabs}
\usepackage{mathtools}
\usepackage{pifont}
\usepackage{calc}

\usepgfplotslibrary{colorbrewer}
\pgfplotsset{cycle list/Dark2}
\usetikzlibrary{positioning,calc,backgrounds}
\usetikzlibrary{decorations.pathreplacing}
\usetikzlibrary{fadings}
\usetikzlibrary{patterns}
\usetikzlibrary{shadows.blur}
\usetikzlibrary{shapes}
\pgfplotsset{compat=1.12,
    /pgfplots/ybar legend/.style={
    /pgfplots/legend image code/.code={%
       \draw[##1,/tikz/.cd,yshift=-0.25em]
        (0cm,0cm) rectangle (3pt,0.8em);},
   },
}
\pgfplotsset{%
  s/.style = {%
	mark=o,mark size=1,thick
  }
}
\pgfplotsset{select coords between index/.style 2 args={
    x filter/.code={
        \ifnum\coordindex<#1\def\pgfmathresult{}\fi
        \ifnum\coordindex>#2\def\pgfmathresult{}\fi
    }
}}
\tikzset{every picture/.style={line width=0.75pt}} 
\definecolor{att}{HTML}{FEE1BA}
\definecolor{pipe}{HTML}{000000}
\definecolor{data}{HTML}{FCE0E1}
\definecolor{pred}{HTML}{CCE7CF}
\definecolor{mlp}{HTML}{C1E7F6}
\definecolor{norm}{HTML}{F1F3C0}
\definecolor{fe}{HTML}{C5BEDF}
\definecolor{fs}{HTML}{FBDFE0}
\definecolor{detb}{HTML}{F2F2F3}
\definecolor{attb}{HTML}{F2F2F3}
\definecolor{preb}{HTML}{F2F2F3}
\definecolor{pool}{HTML}{B22222}
\definecolor{view}{HTML}{FFFFFF}
\tikzstyle{base_node} = [black,draw=black,align=left,anchor=center,rectangle,minimum height=0.75cm,rounded corners=5pt]
\tikzset{xi/.style={base_node,fill=data}}
\tikzset{pred/.style={base_node,fill=pred}}
\tikzset{att/.style={base_node,fill=att}}
\tikzset{mlp/.style={base_node,fill=mlp}}
\tikzset{norm/.style={base_node,fill=norm}}
\tikzset{fe/.style={base_node,fill=fe}}
\tikzset{fs/.style={base_node,fill=fs}}
\tikzset{thinpipe/.style={black,draw=black,rounded corners=3pt}}
\tikzset{pipe/.style={draw=pipe,line width=0.5mm,rounded corners=3pt}}

\usepackage[pagebackref=true,breaklinks=true,colorlinks,bookmarks=false]{hyperref}

\def\cvprPaperID{3822} 
\def\confYear{CVPR 2021}

\begin{document}

\title{Weakly Supervised Action Selection Learning in Video}

\author{
Junwei Ma\thanks{Authors contributed equally to this work.}\\
Layer6 AI\\
{\tt\small jeremy@layer6.ai}
\and
Satya Krishna Gorti\footnotemark[1]\\
Layer6 AI\\
{\tt\small satya@layer6.ai}
\and
Maksims Volkovs\\
Layer6 AI\\
{\tt\small maks@layer6.ai}
\and
Guangwei Yu\\
Layer6 AI\\
{\tt\small guang@layer6.ai}
}

\newcommand{\red}[1]{\textcolor{red}{#1}}
\def \x {x_t}
\def \pbg {a_{t}}
\def \casfg {s_{c,t}}
\def \probfg {s_{c}}
\def \asla {\text{ASL-}{a_t}}
\def \asls {\text{ASL-}{s_{c,t}}}
\newcommand{\bgmodel}[1]{\ensuremath{G_t({#1})}}
\newcommand{\fgmodel}[1]{\ensuremath{F_{c,t}({#1})}}
\def \tpos {\mathcal{T}_{\text{pos}}}
\def \tneg {\mathcal{T}_{\text{neg}}}
\def \tposc {\mathcal{T}^c}
\def \bag {h_{c,t}}
\def \actionness {actionness\xspace}
\newcommand*{\myCDots}{$\cdot$\kern-0.05em$\cdot$\kern-0.05em$\cdot$} 
\def \na {\multicolumn{1}{c}{-}}
\def \nabar {\multicolumn{1}{c|}{-}}
\newcommand{\cmark}{\ding{51}}%
\newcommand{\xmark}{\ding{55}}%

\maketitle

\begin{abstract}
Localizing actions in video is a core task in computer vision. The weakly
supervised temporal localization problem investigates whether this 
task can be adequately solved with only video-level labels, significantly
reducing the amount of expensive and error-prone annotation that is required.
A common approach is to train a frame-level classifier where frames 
with the highest class probability are selected to make a video-level prediction. 
Frame-level activations are then used for localization. However, the absence of frame-level 
annotations cause the classifier to impart class bias on every frame.
To address this, we propose the Action Selection Learning (ASL) approach to capture
the general concept of action, a property we refer to as ``actionness''.
Under ASL, the model is trained with a novel class-agnostic task to 
predict which frames will be selected by the classifier. Empirically, we 
show that ASL outperforms leading baselines on two popular 
benchmarks THUMOS-14 and ActivityNet-1.2, with 10.3\% and 5.7\% relative improvement 
respectively. We further analyze the properties of ASL and demonstrate the
importance of actionness. Full code for this work is available here:
\url{https://github.com/layer6ai-labs/ASL}.
\end{abstract}

\section{Introduction}
Temporal action localization is a fundamental task in computer vision with important
applications in video understanding and modelling.
The weakly supervised localization problem investigates whether this task can be adequately
solved with only video-level labels instead of frame-level annotations. This significantly
reduces the expensive and error-prone labelling required in the fully supervised
setting~\cite{zhao2017slac,shou2018autoloc}, but considerably increases the difficulty of
the problem. A standard approach is to apply multiple instance learning to learn a classifier
over instances, where an instance is typically a frame or a short 
segment~\cite{narayan20193c,paul2018w,lee2020background}.
The classifier is trained using the top-$k$ aggregation over the 
instance class activation sequence to generate video probabilities. Localization is then done by 
leveraging the class activation sequence to generate start and end predictions. 
However, in many cases, the instances that are selected in the top-$k$ contain useful 
information for prediction but not the actual action. Furthermore, with top-$k$ selection 
the classification loss encourages the classifier to ignore action instances that are difficult to
classify. Both of these problems can significantly hamper localization accuracy and stem from 
the general inability of the classifier to capture the intrinsic property of action in instances. 
This property is known as ``actionness'' in the existing literature~\cite{chen2014actionness,luo2015actionness}.

\begin{figure}[t]\centering
\subfloat[Context error\label{fig:fail1}]{
    \includegraphics[width=0.45\linewidth]{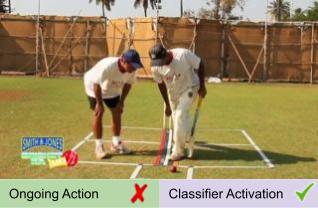}
}
\hfill
\subfloat[Actionness error\label{fig:fail2}]{
    \includegraphics[width=0.45\linewidth]{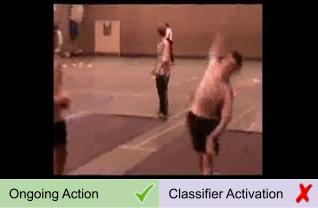}
    }
\vskip -0.2cm
\caption{\protect\subref{fig:fail1} Context error for the ``Cricket Shot'' action due
to the presence of all cricket artifacts but absence of action.
\protect\subref{fig:fail2} Actionness error for the ``Cricket Bowling'' action 
due to the atypical scene despite the presence of action.}
\label{fig:fail}
\vskip -0.4cm
\end{figure}

Ignoring actionness can lead to two major types of error: \emph{context error} and 
\emph{actionness error}. Context error occurs when the classifier activates on instances 
that do not contain actions but contain context indicative of the overall video class~\cite{liu2019completeness,lee2020background}.
Figure~\ref{fig:fail}~\subref{fig:fail1} shows an example of context error. 
Here, cricket players are inspecting a cricket pitch. The instance clearly 
indicates that the video is about cricket and the classifier predicts ``Cricket Shot"
with high confidence. However, no shot happens in this particular instance and 
including it in the localization for ``Cricket Shot" would lead to an error.
Actionness error occurs when the classifier fails to activate on instances
that contain actions. This generally occurs in difficult instances that have 
significant occlusion or uncommon settings. An example of this is shown 
in Figure~\ref{fig:fail}~\subref{fig:fail2}. The action is ``Cricket Bowling", 
but the classifier fails to activate as the scene is indoors and differs from 
the usual cricket setting. 

Leading recent work~\cite{nguyen2019weakly,lee2020background,nguyen2018weakly} in this area
propose an attention model to filter out background and then train a classifier on the
filtered instances to predict class probabilities. This has the drawback of making it more 
challenging for the classifier to learn as important context is potentially removed as background.

Our motivation is to design a learning framework that can use the context information for 
class prediction and at the same time learn to identify action instances for localization. 
We have seen from the supervised setting that leading 
object detection~\cite{girshick2014rich,ren2015faster,cai2018cascade} and temporal
localization~\cite{lin2017single, lin2018bsn, lin2019bmn} methods leverage 
class-agnostic proposal networks to generate highly accurate predictions. This demonstrates that
a general objectness/actionness property can be successfully learned across a diverse set of classes. To this end, we 
propose a new approach called \textbf{A}ction \textbf{S}election \textbf{L}earning (ASL) where the
class-agnostic actionness model learns to predict which frames will be selected in the top-$k$ sets 
by the classifier. During inference, we combine predictions from the actionness model with class
activation sequence and show considerable improvement in localization accuracy. Specifically, 
ASL achieves new state-of-the-art on two popular benchmarks THUMOS-14 and ActivityNet-1.2, where 
we improve over leading baselines by 10.3\% and 5.7\% in mAP respectively. We further analyze the 
performance of our model and demonstrate the advantages of the actionness approach.

\section{Related Work}

\textbf{Weakly Supervised Temporal Action Localization } A prominent direction in the 
weakly supervised setting is to leverage the class activation sequence to improve localization.
UntrimmedNet~\cite{wang2017untrimmednets} focuses on improving the instance selection 
step using class activations. Hide-and-seek~\cite{singh2017hide} applies instance dropout to reduce 
classifier's dependence on specific instances. W-TALC~\cite{paul2018w} 
incorporates a co-activity similarity loss to capture inter-class and inter-video relationships.
3C-Net~\cite{narayan20193c} adopts a center loss to reduce inter-class variations while applying 
additional action-count information for supervision. Focusing on class activations
can be susceptible to context error, and a parallel line of research explores how to 
identify context instances. STPN~\cite{nguyen2018weakly} extends UntrimmedNet by introducing a 
class-agnostic attention model with sparsity constraints. BM~\cite{nguyen2018weakly} 
uses self-attention to separate action and context instances. CMCS~\cite{liu2019completeness} 
assumes a stationary prior on context and leverages it to model context instances. 
BaSNet~\cite{lee2020background} explicitly models a separate context class that is used to filter 
instances during inference. DGAM~\cite{shi2020weakly} trains a variational autoencoder 
to model the class-agnostic instance distribution conditioned on attention to separate 
context from actions. More recently, TSCN~\cite{zhaitwo} and EM-MIL~\cite{luo2020weakly} 
propose two-stream architectures. TSCN separates the RGB and Flow modules and learns from 
pseudo labels generated by combining the predictions of the two streams. EM-MIL introduces 
a key instance and a classification module trained alternately to maintain the multi 
instance learning assumption.

\textbf{Actionness Learning } Our approach is motivated by related work in the supervised 
setting where a common design choice is to learn a class-agnostic module 
to generate proposals that are then labelled by the classifier~\cite{lin2017single,lin2018bsn,lin2019bmn}. 
Earlier work defines actionness as a likelihood of a generic but deliberate action 
that is separate from context~\cite{chen2014actionness}, and applies it to detect human activity 
in both image~\cite{chen2014actionness} and video~\cite{yu2015fast} settings. A related concept 
of ``interestingness'' has been proposed to identify actions at the pixel level~\cite{wang2016actionness}.
Work in action recognition shows that generic attributes exist across action classes and can 
be leveraged for recognition~\cite{luo2015actionness}. Similar concept has been demonstrated 
to be successful for tracking applications~\cite{li2016searching}. Finally, in object detection, leading approaches heavily leverage class-agnostic proposal networks to 
first identify regions of high ``objectness''~\cite{girshick2014rich,ren2015faster,cai2018cascade}. 

In this work, we demonstrate that the analogous ``actionness'' property in videos can be effectively learned with only video-level labels.
\section{Approach}

We treat a video as a set of $T$ instances
$\{x_1,...,x_T\}$, dropping video index to reduce notation clutter.
An instance can be a frame or a fixed-interval segment,
represented by a feature vector $x_t\in\mathbb{R}^d$.
In the weakly supervised temporal localization task, each instance $x_t$ 
either contains an action from one of $C$ classes or is the background, however, 
this is unknown to us. Instead, we are given video-level classes $Y\subseteq\{1,...,C\}$ 
which is the union of all instance classes in the video.
The weakly supervised temporal localization task then asks whether 
video-level class information can be used to localize actions
across instances. In this section, we first outline the classification 
framework in Section~\ref{sec:base}, and then describe our approach 
in Section~\ref{sec:asl}.

\subsection{Base Classifier}\label{sec:base}

We define a video classifier to predict target video-level classes as:
\begin{align}\label{eq:casfg}
    \casfg &= \fgmodel{x_1,...,x_T}
\end{align}
where $F$ is a neural network applied to the entire video, and $\fgmodel{\cdot}$ denotes its
output at class $c$ and instance $x_t$. Taken over all $T$ instances we refer to 
$\fgmodel{\cdot}$ as the class activation sequence (CAS).
Multiple instance learning~\cite{carbonneau2018multiple} is commonly used
to train the classifier, where top-$k$ pooling is applied over CAS for each class to aggregate
the highest activated instances and make video-level predictions. We denote the set of top-$k$
instances for each class as $\tposc$ :
\begin{align}\label{eq:tposc}
    \tposc &= \underset{\substack{\mathcal{T}\subseteq\{1,...,T\}\\|\mathcal{T}|=k}}{\arg\max}
    \sum_{t\in\mathcal{T}} \bag 
\end{align} 
where $k$ is a hyper-parameter and $\bag$ is the instance selection probability
that is used to select the top instances. In prior work, the selection probability 
is straightforwardly set to the CAS with $\bag=\casfg$. However, we make a deliberate distinction here
which allows incorporating actionness as we demonstrate in the following section.
Aggregation, such as mean pooling, is applied over the selected instances in $\tposc$
to make video-level class prediction:
\begin{equation}\label{eq:fgprob}
     p_{c} = \text{softmax}\left(\frac{1}{|\tposc|}\sum_{t\in\tposc}\casfg \right)
\end{equation}
Finally, this model is optimized with the multiple instance learning objective:
\begin{equation}\label{eq:Lcls}
    \mathcal{L}_{\text{CLS}} = -\frac{1}{|Y|}\sum_{c\in Y} \log p_{c} 
\end{equation}\textbf{\textbf{}}

\subsection{Action Selection Learning in Video}\label{sec:asl}

The classifier introduced in the previous section optimizes the classification 
objective which encourages only instances that strongly support the target video
classes to get selected in the top-$k$ set. This can lead to the inclusion of instances that provide strong 
context support but do not contain the action (actionness error), and also the exclusion of instances
that contain the action but are difficult to predict (context error). Both of these problems
do not affect video classification accuracy, but can significantly hurt 
localization. We address this by developing a novel action selection learning (ASL)
approach to capture the class-agnostic actionness property of each instance.
The main idea behind ASL is that the top-$k$ set $\tposc$ used for prediction 
is likely capturing both context and action instances. However, context information
is highly class-specific whereas actions share similar characteristics across classes.
Consequently, by training a separate class-agnostic model to predict whether an instance will be in the 
top-$k$ set for any class, we can effectively capture instances that contain actions and
filter out context. We begin by defining a neural network \actionness model $G$:
\begin{align}
    \pbg &= \sigma\left(\bgmodel{x_1,...,x_T}\right) \label{eq:pbg}
\end{align}
where $\sigma$ is the sigmoid activation function, and $\bgmodel{\cdot}$ denotes 
the output of $G$ for instance $x_t$. Here, $\pbg$ can be interpreted as the 
probability that $x_t$ contains any action.
\begin{figure*}[tp!]
    \centering
    \subfloat[Diagram of the proposed approach. \label{fig:arch}]{
        \centering
        \resizebox{0.46\textwidth}{!}{
                    \begin{tikzpicture}[align=center,node distance=1 and 1]
                        \def \dx {1}
                        \def \dy {2}
                        
                        \node[xi] (x) at (0,0) {$x_t$};
                        \node[xi,above right = of x] (s) {$\casfg$};
                        \node[xi,below right = of x] (a) {$\pbg$};
                        \node[xi,above right = of a] (h) {$\bag$};
                        \node[fs,right = of h] (tposc) {$\tposc$};
                        \node[fs, right = of tposc] (tpos) {$\tpos$};
                        \node[fs, right = of tpos] (tneg) {$\tneg$};
                        \node[xi, above = of tposc] (ps) {$p_c$};
                        \node[pred, right = of ps] (lcls) {$\mathcal{L}_{\text{CLS}}$};
                        \node[pred, below = of tpos] (lact) {$\mathcal{L}_{\text{ASL}}$};
                        \node[fe, right = of lcls] (y) {$Y$};
                        \draw[->] (x) -- (s) node [pos=0.9,fill=none,anchor=south east] {$F_{c,t}$};
                        \draw[->] (x) -- (a) node [midway,fill=none,anchor=north east] {$G_t$};
                        \draw[->] (a) -- (h);
                        \draw[->] (s) -- (h);
                        \draw[->] (h) -- (tposc);
                        \draw[->] (tposc) -- (tpos);
                        \draw[->] (y) -- (tpos);
                        \draw[->] (tpos) -- (tneg);
                        \draw[->] (tposc) -- (ps);
                        \draw[->] (s)  -- (ps);
                        \draw[->] (a)  -- (lact);
                        
                        \begin{scope}[on background layer]
                            \draw[rounded corners=7,fill=yellow, fill opacity=0.1] (-0.5,2.6) rectangle (4,-2.3);
                            \draw[rounded corners=7,fill=blue, fill opacity=0.1] (0.5,3) rectangle (5.75,-0.6);
                            \node[anchor=north east] at (4,2.6) {\LARGE$T$};
                            \node[anchor=north east] at (5.75,3) {\LARGE$C$};
                        \end{scope}
                        
                        \draw[->,dashed] (tpos) -- (lact);
                        \draw[->,dashed] (tneg) -- (lact);
                        
                        \draw[->] (ps) -- (lcls);
                        \draw[->,dashed] (y) -- (lcls);
                    \end{tikzpicture}
            }
        }%
        \hspace{0.9cm}
    \subfloat[Toy example with $C=4,T=7,k=3$.\label{fig:action_select}]{
        \centering
        \resizebox{0.46\textwidth}{!}{
            \begin{tikzpicture}[x=0.75pt,y=0.75pt,yscale=-1,xscale=1]

\definecolor{node0}{HTML}{FFFFFF}
\definecolor{node1}{HTML}{F1DFB4}
\definecolor{nodepos}{HTML}{B8E986}
\definecolor{nodeneg}{HTML}{FFFFFF}
\definecolor{tpos}{HTML}{B4D0F1}
\definecolor{tneg}{HTML}{EBBBAE}
\definecolor{act}{HTML}{CFCFCF}

\begin{scope}[shift={(0,0)}]
\draw  [fill=node0  ,fill opacity=1 ] (350,30) -- (370,30) -- (370,50) -- (350,50) -- cycle ;
\draw  [fill=node0  ,fill opacity=1 ] (390,30) -- (410,30) -- (410,50) -- (390,50) -- cycle ;
\draw  [fill=node0  ,fill opacity=1 ] (470,30) -- (490,30) -- (490,50) -- (470,50) -- cycle ;
\draw  [fill=node0  ,fill opacity=1 ] (350,50) -- (370,50) -- (370,70) -- (350,70) -- cycle ;
\draw  [fill=node0  ,fill opacity=1 ] (430,50) -- (450,50) -- (450,70) -- (430,70) -- cycle ;
\draw  [fill=node0  ,fill opacity=1 ] (450,50) -- (470,50) -- (470,70) -- (450,70) -- cycle ;
\draw  [fill=node0  ,fill opacity=1 ] (470,50) -- (490,50) -- (490,70) -- (470,70) -- cycle ;
\draw  [fill=node0  ,fill opacity=1 ] (430,70) -- (450,70) -- (450,90) -- (430,90) -- cycle ;
\draw  [fill=node1  ,fill opacity=1 ] (370,70) -- (390,70) -- (390,90) -- (370,90) -- cycle ;
\draw  [fill=node1  ,fill opacity=1 ] (390,70) -- (410,70) -- (410,90) -- (390,90) -- cycle ;
\draw  [fill=node1  ,fill opacity=1 ] (350,70) -- (370,70) -- (370,90) -- (350,90) -- cycle ;
\draw  [fill=node0  ,fill opacity=1 ] (410,70) -- (430,70) -- (430,90) -- (410,90) -- cycle ;
\draw  [fill=node1  ,fill opacity=1 ] (410,50) -- (430,50) -- (430,70) -- (410,70) -- cycle ;
\draw  [fill=node1  ,fill opacity=1 ] (370,50) -- (390,50) -- (390,70) -- (370,70) -- cycle ;
\draw  [fill=node1  ,fill opacity=1 ] (390,50) -- (410,50) -- (410,70) -- (390,70) -- cycle ;
\draw  [fill=node0  ,fill opacity=1 ] (450,70) -- (470,70) -- (470,90) -- (450,90) -- cycle ;
\draw  [fill=node0  ,fill opacity=1 ] (470,70) -- (490,70) -- (490,90) -- (470,90) -- cycle ;
\draw  [fill=node1  ,fill opacity=1 ] (410,30) -- (430,30) -- (430,50) -- (410,50) -- cycle ;
\draw  [fill=node1  ,fill opacity=1 ] (430,30) -- (450,30) -- (450,50) -- (430,50) -- cycle ;
\draw  [fill=node0  ,fill opacity=1 ] (450,30) -- (470,30) -- (470,50) -- (450,50) -- cycle ;
\draw  [fill=node1  ,fill opacity=1 ] (370,30) -- (390,30) -- (390,50) -- (370,50) -- cycle ;

\draw  [fill=node0  ,fill opacity=1 ] (350,30) -- (370,30) -- (370,10) -- (350,10) -- cycle ;
\draw  [fill=node0  ,fill opacity=1 ] (370,30) -- (390,30) -- (390,10) -- (370,10) -- cycle ;
\draw  [fill=node1  ,fill opacity=1 ] (390,30) -- (410,30) -- (410,10) -- (390,10) -- cycle ;
\draw  [fill=node0  ,fill opacity=1 ] (410,30) -- (430,30) -- (430,10) -- (410,10) -- cycle ;
\draw  [fill=node1  ,fill opacity=1 ] (430,30) -- (450,30) -- (450,10) -- (430,10) -- cycle ;
\draw  [fill=node0  ,fill opacity=1 ] (450,30) -- (470,30) -- (470,10) -- (450,10) -- cycle ;
\draw  [fill=node1  ,fill opacity=1 ] (470,30) -- (490,30) -- (490,10) -- (470,10) -- cycle ;

\draw  [fill=tpos  ,fill opacity=1 ] (370,110) -- (390,110) -- (390,130) -- (370,130) -- cycle ;
\draw  [fill=tpos  ,fill opacity=1 ] (390,110) -- (410,110) -- (410,130) -- (390,130) -- cycle ;
\draw  [fill=tpos  ,fill opacity=1 ] (350,110) -- (370,110) -- (370,130) -- (350,130) -- cycle ;
\draw  [fill=tpos  ,fill opacity=1 ] (410,110) -- (430,110) -- (430,130) -- (410,130) -- cycle ;
\draw  [fill=tneg  ,fill opacity=1 ] (450,110) -- (470,110) -- (470,130) -- (450,130) -- cycle ;
\draw  [fill=tneg  ,fill opacity=1 ] (470,110) -- (490,110) -- (490,130) -- (470,130) -- cycle ;
\draw  [fill=tneg  ,fill opacity=1 ] (430,110) -- (450,110) -- (450,130) -- (430,130) -- cycle ;

\begin{scope}[shift={(10,0)}]
    \draw  [fill=nodeneg  ,fill opacity=1 ] (510,50) -- (530,50) -- (530,70) -- (510,70) -- cycle ;
    \draw  [fill=nodeneg  ,fill opacity=1 ] (510,70) -- (530,70) -- (530,90) -- (510,90) -- cycle ;
    \draw  [fill=nodeneg  ,fill opacity=1 ] (510,30) -- (530,30) -- (530,50) -- (510,50) -- cycle ;
    \draw  [fill=nodeneg  ,fill opacity=1 ] (510,30) -- (530,30) -- (530,10) -- (510,10) -- cycle ;
    \node[anchor=center] at (520,20) {$0$};
    \node[anchor=center] at (520,40) {$0$};
    \node[anchor=center] at (520,60) {$1$};
    \node[anchor=center] at (520,80) {$1$};
    \draw (520,00) node [anchor=center][inner sep=0.75pt]   [align=left] {$\displaystyle Y$};
\end{scope}

    \draw [decorate,decoration={brace,amplitude=5pt},xshift=0pt,yshift=-3]
    (350,10) -- (490,10) node [black,midway,yshift=12] 
    {\footnotesize $T$};

    \draw [decorate,decoration={brace,amplitude=5pt},xshift=3,yshift=0]
    (490,10) -- (490,90) node [black,midway,xshift=12] 
    {\footnotesize $C$};
    
    \draw [decorate,decoration={brace,amplitude=5pt,mirror},xshift=0,yshift=3]
    (350,130) -- (430,130) node [black,midway,yshift=-12] 
    {\footnotesize $ \mathcal{T}_{\text{pos}}$};
    \draw [decorate,decoration={brace,amplitude=5pt,mirror},xshift=0,yshift=3]
    (430,130) -- (490,130) node [black,midway,yshift=-12] 
    {\footnotesize $ \mathcal{T}_{\text{neg}}$};
    

    
\end{scope}

\begin{scope}[shift={(30,-10)}]
    \draw [->] (225,60) -- (240,60) ;
    \draw [->] (300,60) -- (315,30) ;
    \draw [->] (300,60) -- (315,50) ;
    \draw [->] (300,60) -- (315,70) ;
    \draw [->] (300,60) -- (315,90) ;
    \draw [->] (225,120) -| (260,120) |- (260,75);
    \draw (270,60) node [anchor=center][inner sep=0.75pt]   [align=left] {$\displaystyle h(a_{t} ,s_{c,t})$};
    
    \draw (220,60) node [xi,anchor=east]   [align=left] {$\displaystyle s_{c,t}$};
    \draw (220,120) node [xi,anchor=east]   [align=left] {$\displaystyle a_{t}$};
    
\end{scope}


\end{tikzpicture}
        }
        }
        \caption{ASL model architecture and toy example.
        \protect\subref{fig:arch} For every instance $x_t$ the classifier $F_{c,t}$ predicts
        class activation $s_{c,t}$, and actionness model $G_t$ predicts actionness score
        $a_t$. Class activation and actionness are combined with the instance selection function $h$
        to get instance selection probability $h_{c,t} = h(a_{t}, s_{c,t})$. Top-$k$ instances 
        with the highest selection probabilities are then added to $\tposc$ and aggregated 
        together to generate class prediction $p_c$ for the video. Finally, the union of top-$k$ instances 
        across ground-truth classes $Y$ is used to generate target sets $\tpos$ and $\tneg$ for the actionness
        model.
        \protect\subref{fig:action_select}  Toy example illustrates how target sets $\tpos$ and $\tneg$
        are computed. The video has $T=7$ instances, $C=4$ classes and $k = 3$. 
        For each class we select top-3 instances with the highest action selection probabilities 
        $h(a_{t}, s_{c,t})$ indicated by yellow cells. Taking union of selected instances across 
        ground truth classes ($c \in Y$) we get $\tpos$ shown in blue. All other instances form
        $\tneg$ shown in red.}
        \label{fig:diagram}
        \vspace{-1.5em}
\end{figure*}
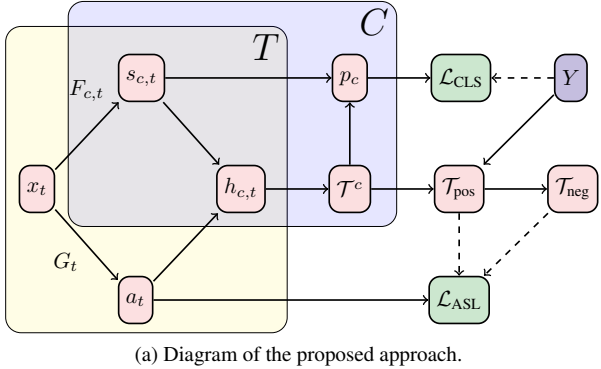
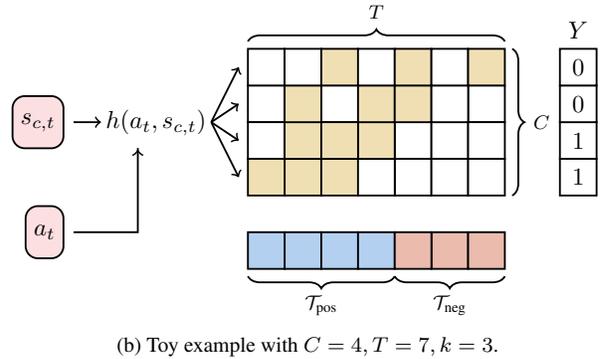

For an instance to contain a specific action, it should simultaneously contain
evidence of the corresponding class and evidence of actionness. As we discussed,
class evidence alone is not sufficient and can lead to context and 
actionness errors. To account for both properties, we expand the instance 
selection function:
\begin{align}\label{eq:bagselect}
   \bag = h(\pbg, \casfg) 
\end{align}
This selection function combines beliefs from both models and can be implemented in
multiple ways. In this work we fuse the scores with a convex combination
$h(\pbg,\casfg) = \beta \pbg + (1-\beta) \casfg$ and leave other possible 
architectures for future work. After $\bag$ is computed, we proceed as before 
and select top-$k$ instances with the highest $\bag$ values to get $\tposc$.

To train the \actionness model $G$, we design a novel task to
predict whether a given instance $x_t$ will be in the top-$k$ set for any 
ground truth class. Since context is highly class dependent, we hypothesize that 
$G$ can only perform well on this task by learning to capture
action characteristics that are ubiquitous across classes. This hypothesis is 
further motivated by the fact that many leading supervised localization methods first 
generate class-agnostic proposals and then predict classes for 
them~\cite{lin2017single, lin2018bsn, lin2019bmn}. High accuracy of these models
indicates that the proposal network is able to learn the general actionness 
property independent of the class, and we aim to do the same here. We first 
partition instances into positive and negative sets:
\begin{align}
    \tpos&=\bigcup_{c\in Y}\tposc \label{eq:tpos}\\
    \tneg&=\{1,...,T\} \backslash \tpos \label{eq:tneg}
\end{align}
where positive set $\tpos$ contains the union of all instances that appear
in the top-$k$ for the ground truth classes $Y$, and negative set $\tneg$ has all other 
instances. We then train $G$ to predict whether each instance is in the 
positive or negative set. In our model the classifier and \actionness networks are tied 
by the instance selection function. Empirically, we observe that during training
as classification accuracy improves better instances get selected into positive 
and negative sets. This improves the \actionness model which translates to better 
top-$k$ instance selection for the classifier, leading to further improvement 
in classification accuracy. The two models are thus complementary to each other, 
and we show that {\it both} classification and localization accuracy improve 
when the \actionness network is added.

Since our target positive and negative sets contain both context and
action instances, the binary inclusion labels can be noisy. This is particularly 
the case early in training when classification accuracy is poor and
top selected instances are not accurate. Traditional cross entropy 
classification loss assigns a large penalty when prediction deviates significantly from the ground truth.
This is a desirable property when labels are clean, enabling the
model to converge quickly~\cite{zhang2018generalized}. However, recent
work shows that cross entropy leads to poor performance under noisy labels, 
where the high penalty can force the model to overfit to noise~\cite{ghosh2017robust,zhang2018generalized}.
To address this problem the generalized cross entropy loss has been 
proposed that softens the penalty in regions of high disagreement~\cite{zhang2018generalized}. 
We adopt this loss here to improve the generalization of the actionness model:
\begin{equation}\label{eq:Lact}\small
    \mathcal{L}_{\text{ASL}} = \frac{1}{|\tpos|} \sum_{t\in\tpos} \frac{1-(\pbg)^q}{q}
    + \frac{1}{|\tneg|} \sum_{t\in\tneg}\frac{1-(1-\pbg)^q}{q} 
\end{equation}
where $0 < q \leq 1$ controls the noise tolerance. $\mathcal{L}_{\text{ASL}}$ is based on 
the negative Box-Cox transform~\cite{box1964analysis}, and approaches mean absolute
error when $q$ is close to 1 which is more tolerant to deviations from ground truth. 
On the other hand, as $q$ approaches 0, $\mathcal{L}_{\text{ASL}}$ behaves similarly to the 
cross entropy loss with stronger penalties. By appropriately setting $q$ we can control model 
sensitivity to noise and improve robustness. During training we optimize both 
classification and ASL losses simultaneously $\mathcal{L} = \mathcal{L}_{\text{CLS}} + \mathcal{L}_{\text{ASL}}$ 
and backpropagate the gradients through both classifier and actionness networks.

The proposed ASL architecture is summarized in Figure~\ref{fig:diagram}\subref{fig:arch}.
Figure~\ref{fig:diagram}\subref{fig:action_select} also shows a toy example 
that illustrates how positive and negative sets $\tpos$ and $\tneg$ are computed.
The video has $T=7$ instances and $C=4$ classes, two of which are in the 
ground truth $Y = \{3, 4\}$. Moreover, $k=3$ so for each class top-3 instances with the
highest instances selection probabilities $h(a_{t}, s_{c,t})$ are selected, 
indicated by yellow cells. Union of instances selected for the ground truth
classes form $\tpos = \{x_1, x_2, x_3, x_4\}$ shown in red, and all 
other instances form $\tneg = \{x_5, x_6, x_7\}$ shown in blue. To successfully 
predict instances in each list, the actionness model must find commonalities 
between all instances in $\tpos$ and distinguish them from $\tneg$. As we 
demonstrate in the experimental section this commonality is the 
presence of actionness which significantly aids the localization task.

After training, we use the instance selection probabilities $h_{c,t}$ to localize 
actions in test videos. Given a test video with $T'$ instances, we run it 
through our model to get the corresponding instance selection probability 
sequence $h_{c,1}, ..., h_{c,T'}$. We then follow recent work~\cite{narayan20193c,lee2020background,shi2020weakly}  
and apply multiple thresholds $0 < \alpha < 1$. All instances where selection
probability is above the threshold  $h_{c,t} > \alpha$ are considered selected, 
and we take all consecutive sequences as proposal candidates. Repeating 
this process for each threshold, we obtain a set of proposals for each class.
We then apply non-maximal suppression to eliminate overlapping and similar 
proposals and generate the final localization predictions.
\begin{table}[tp!]
\sisetup{round-mode=places,round-precision=1}
\small
\centering
\caption{THUMOS-14 results. mAP is the mean of AP@IoU scores across thresholds $\{0.1, 0.2,...,0.9\}$.
Ablation results are shown at the bottom where $\asls$ and $\asla$ use class
probability $s_{c,t}$ and actionness probability $a_t$ respectively to localize, 
and ASL-BCE trains the actionness network $G$ with the binary cross entropy loss.}
\vskip -0.2cm

\def \nabarz {\text{-}}
\newcolumntype{C}[1]{>{\centering\arraybackslash\leavevmode}m{#1}}
\newcolumntype{L}[1]{>{\raggedright\arraybackslash}m{#1}}

\begin{tabular}{l | c H c H c H c H c | c}
\hline
\multirow{2}{*}{Approach}& \multicolumn{9}{c|}{AP@IoU} & \multirow{2}{*}{mAP}\\
 & 0.1 & 0.2 & 0.3 & 0.4 & 0.5 & 0.6 & 0.7 & 0.8 & 0.9 & \\
\hline

CMCS~\cite{liu2019completeness} & 57.4 & 50.8 & 41.2 & 32.1 & 23.1 & 15.0 & 7.0 & \nabarz & \nabarz & \nabarz\\
MAAN~\cite{yuan2019marginalized} & 59.8 & 50.8 & 41.1 & 30.6 & 20.3 & 12.0 & 6.9 & 2.6 & 0.2 & 24.9 \\
3C-Net~\cite{narayan20193c} & 56.8 & 49.8 & 40.9 & 32.3 & 24.6 & \nabarz & 7.7 & \nabarz & \nabarz & \nabarz \\
BaSNet~\cite{lee2020background}&58.2&52.3&44.6&36.0&27.0&18.6&10.4&3.9&0.5& 27.9\\
BM~\cite{nguyen2019weakly}&60.4&56.0&46.6&37.5&26.8&17.6&9.0&3.3&0.4& 28.6\\
DGAM~\cite{shi2020weakly} & 60.0 & 54.2 & 46.8 & 38.2 & 28.8 & 19.8 & 11.4 & 3.6 & 0.4& 29.2 \\
ACL~\cite{Gong_2020_CVPR} & \nabarz & \nabarz & 46.9 & 38.9 & 30.1 & 19.8 & 10.4 & \nabarz & \nabarz & \nabarz \\
TSCN~\cite{zhaitwo}& 63.4 & 57.6 & 47.8 & 37.7 & 28.7 & 19.4 & 10.2 & 3.9 & 0.7 & 22.9 \\
EM-MIL~\cite{luo2020weakly} & 59.1 & 52.7 & 45.5 & 36.8 & 30.5 & \textbf{22.7} & \textbf{16.4} & \nabarz & \nabarz  & \na \\

\textbf{ASL (ours)} & \textbf{67.0} & \textbf{61.0} & \textbf{51.8} & \textbf{42.0}& \textbf{31.1} & 20.1 & 11.4 & \textbf{4.2} & \textbf{0.7} & \textbf{32.2} \\

\hline
\multicolumn{11}{c}{Ablation}\\
\hline

$\asls$ & 56.9 & 48.7 & 40.5 & 29.9 & 19.7 & 11.7 & 6.0 & 2.2 & 0.4 & 24.0\\

$\asla$ & 55.9 & 48.7 & 40.3 & 30.3 & 20.6 & 12.8 & 6.8 & 2.5 & 0.4 & 30.4\\ 

ASL-BCE & 66.4 & 59.9 & 50.5 & 41.0 & 30.5 & 19.8 & 10.9 & 4.3 & 0.7 & 31.6\\

\hline

\end{tabular}
\label{tab:thumos}
\vspace{-0.7em}
\end{table}

\section{Experiments}

We conduct extensive experiments on two popular weakly supervised temporal localization
datasets containing untrimmed videos: THUMOS-14~\cite{jiang2014thumos} and 
ActivityNet-1.2~\cite{caba2015activitynet}. THUMOS-14 contains 200 training videos with 20 action classes
and 212 test videos. ActivityNet-1.2 contains 4,819 training and 2,383 test videos with 100 action classes.
Both datasets have videos that vary significantly in length from a few seconds to over 25 minutes.
This makes the problem challenging since the model has to perform well on both long and short action sequences.
For all experiments, we only use video-level class labels during training. To make the comparison fair,
we follow the same experimental setup used in literature~\cite{paul2018w,narayan20193c,lee2020background,shi2020weakly},
including data splits, evaluation metrics and input features. 
For all experiments, we report average precision (AP) at the different intersection over union (IoU) 
thresholds between predicted and ground truth localizations. For brevity, we show selected thresholds in the results table for both datasets. The mAP is computed with IoU thresholds between 0.1 to 0.9 with increments of 0.1 on THUMOS-14 and between 0.5 to 0.95 with increments of 0.05 on ActivityNet-1.2
to stay consistent with previous work. Full results on all thresholds used for computing mAP are found in the supplementary material.

{\bf Implementation Details } We generate instance input features $x_t$ following the
same pipeline as recent leading approaches~\cite{narayan20193c,lee2020background,shi2020weakly}.
The I3D network~\cite{carreira2017quo} pre-trained on the Kinetics dataset~\cite{kay2017kinetics} 
is applied on each sub-sequence of 16 consecutive frames with RGB and TVL1 flow~\cite{wedel2009improved} 
inputs to extract 2048-dimensional feature representation by spatiotemporally pooling the Mixed5c layer.
Linear interpolation across time is then applied for both datasets. To make the comparison fair 
we adopt the same base classifier for $F$ as in~\cite{lee2020background} with 512 hidden 
units and ReLU activations. Similarly, the actionness network $G$ is fully connected with
512 hidden units. Both networks are applied across time to every instance and 
operate similarly to convolutional filters with kernel size 1. We set noise tolerance 
$q = 0.7$ for both datasets, and use previously reported instance selection parameters $k = T/8$ 
for THUMOS-14 and ActivityNet-1.2~\cite{narayan20193c,lee2020background}. 
We set $\beta$ to be 0.5 for the instance selection function $h(\pbg, \casfg)$. ASL is trained 
using the ADAM optimizer~\cite{kingma2014adam} with batch size 16, learning rate 1e-4 
and weight decay 1e-4 until convergence. During inference, we use ten localization 
thresholds $\alpha$ from 0 to 1 to generate localization proposals. We then compute 
final predictions by applying non-maximum suppression to eliminate overlapping and similar proposals.
\begin{table}[tp!]\small\centering
\caption{ActivityNet-1.2 results. mAP is the mean of AP@IoU scores across thresholds $\{0.5, 0.55,...,0.95\}$.}
\vskip -0.2cm

    }
    \end{centering}
    \vspace{-0.0cm}
    \caption{THUMOS-14 per-class AP results, four largest relative improvements are highlighted in blue.}\label{fig:ablate_class}
    \vspace{-0.4cm}
\end{figure*}

\begin{figure*}[t]
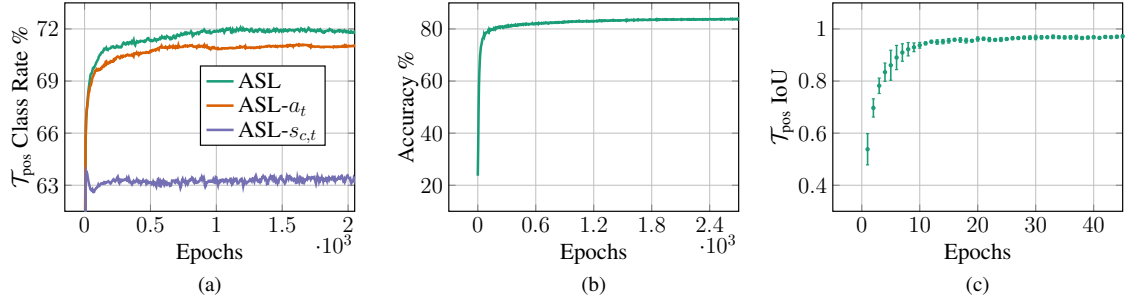
\centering
    \newcommand{\fabwidth}{.24\linewidth}
    \pgfplotsset{
        width=8cm, height=6.2cm,
    }
    \setlength\tabcolsep{13pt} 
    %
    \vskip -0.3cm
    \caption{
        THUMOS-14 $\tpos$ analysis.
        \protect\subref{fig:posk} Fraction of instances in $\tpos$ that contain ground truth action from any target class.
        \protect\subref{fig:aslbce} Validation ASL accuracy at predicting which instances will be in $\tpos$.
        \protect\subref{fig:iou} Averaged across training videos IoU of $\tpos$ sets between consecutive epochs, error bars show one standard deviation.
        }
        \label{fig:ablation}
        \vspace{-0.2cm}
\end{figure*}

We compare our approach with an extensive set of leading recent baselines: TSM~\cite{yu2019temporal}, CMCS~\cite{liu2019completeness}, 
MAAN~\cite{yuan2019marginalized}, 3C-Net~\cite{narayan20193c}, CleanNet~\cite{liu2019weakly}, BaSNet~\cite{lee2020background}, 
BM~\cite{nguyen2019weakly}, DGAM~\cite{shi2020weakly}, TSCN~\cite{zhaitwo} and EM-MIL~\cite{luo2020weakly}. Details 
for each baseline can be found in the related work section, and we directly use the results reported by the 
respective authors.
\setlength\fboxrule{0pt}
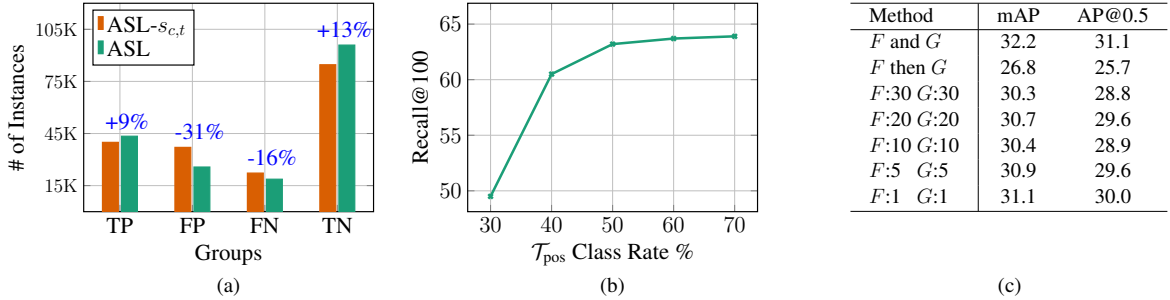
\begin{figure*}[t]\centering
    \newcommand{\fabwidth}{.24\linewidth}
    \pgfplotsset{
        width=8cm, height=6.2cm,
    }
    \setlength\tabcolsep{13pt} 
    \begin{tabular}{ccc}
            \resizebox{\fabwidth}{!}{
                \centering
                \fbox{
                    \begin{tikzpicture}[trim axis right,trim axis left,baseline]
                    \begin{axis}[
                        grid=major,
                        ybar,
                        ymin=0,ymax=120,ytick={15,45,75,105},
                        yticklabels={
                        $15\mathrm{K}$,$45\mathrm{K}$,$75\mathrm{K}$,$105\mathrm{K}$
                        },
                        xtick pos=left,
                        xmin=-0.5,xmax=3.5,
                        yticklabel style={
                            font=\large,
                        },
                        legend cell align={left},
                        legend pos=north west,
                        xtick=data,
                        xticklabels={
                        TP,FN,FP,TN
                            },
                        font=\Large,
                        tick align=inside,
                        xlabel={Groups},
                        ylabel={\# of Instances},
                        every axis plot/.append style={fill},
                        ]
                    \addplot[index of colormap=1 of Dark2,y filter/.code={\pgfmathparse{\pgfmathresult/1000}\pgfmathresult}] coordinates {(0,39914)(2,22221)(1,37019)(3,84596) };
                    \addplot[index of colormap=0 of Dark2,y filter/.code={\pgfmathparse{\pgfmathresult/1000}\pgfmathresult}] coordinates {(0,43442)(2,18693)(1,25704)(3,95911) };
                    \node[] at (axis cs: 0.1,52) {\textcolor{blue}{+9\%}};
                    \node[] at (axis cs: 2.1,31) {\textcolor{blue}{-16\%}};
                    \node[] at (axis cs: 1.1,46) {\textcolor{blue}{-31\%}};
                    \node[] at (axis cs: 3.1,105) {\textcolor{blue}{+13\%}};
                    
                    \legend{$\asls$,ASL}
                    \end{axis}
                    \end{tikzpicture}
                }
            }
        &
            \resizebox{\fabwidth}{!}{
                \centering
                \fbox{
                    \begin{tikzpicture}[trim axis right,trim axis left,baseline]
                      \begin{axis}[
    grid=major,
    font=\Large,
    xmin=26.25,xmax=73.75,
    ymin=48.125,ymax=66.875,
    xlabel={\smash{$\tpos$} Class Rate \%},
    ylabel={Recall@100},
    legend cell align={left},
    legend style={
        at={(0.9,0.7)},
        anchor=north east,
    }
    ]
\addplot+[mark=x,ultra thick] coordinates {
(70, 63.9)
(60, 63.7)
(50, 63.2)
(40, 60.5)
(30, 49.5)
};
                      \end{axis}
                    \end{tikzpicture}
                }
            }
        &
            \resizebox{.26\linewidth}{!}{
            \centering\fbox{
 \Large
 \begin{tabular}[b]{l| c c}
     \hline
     Method & mAP & AP@0.5  \\
     \hline
     $F$ and $G$  & 32.2 & 31.1 \\
     $F$ then $G$  & 26.8 & 25.7 \\
     $F$:30 $G$:30 & 30.3 & 28.8 \\
     $F$:20 $G$:20 & 30.7 & 29.6 \\
     $F$:10 $G$:10 & 30.4 & 28.9 \\
     $F$:5 \ \ $G$:5 & 30.9 & 29.6 \\
     $F$:1 \ \ $G$:1 & 31.1 & 30.0 \\
     \hline
\end{tabular}
            }
            }
        \\[-0.6cm]
        \fbox{\subfloat[\label{fig:error}]{\quad}}   &
        \fbox{\subfloat[\label{fig:recall}]{\quad}}   &
        \fbox{\subfloat[\label{tab:fgschedule}]{\quad}}   \\
    \end{tabular}
    \vskip -0.3cm
    \caption{
        THUMOS-14 error and ablation analysis.
        \protect\subref{fig:error} Error analysis across all test instances by
        type. TP, FP, TN, FN indicate true positives, false positives, true negatives and false negatives respectively.
        \protect\subref{fig:recall}  Class-agnostic localization Recall@100
        using $G$ when it is trained from scratch on $\tpos$ with different
        proportion of instances that contain action.
        \protect\subref{tab:fgschedule} Effect of training schedules in ASL. $F$
        and $G$ is joint ASL training, $F$ then $G$ is sequential training of $F$
        followed by $G$, $F$:$n$ $G$:$m$ is an alternating schedule where $F$ is
        updated for $n$ epochs followed by $G$ for $m$ epochs.
        }
        \label{fig:ablate2}
        \vspace{-0.2cm}
\end{figure*}

{\bf Results } Table~\ref{tab:thumos} summarizes temporal localization results on the THUMOS-14 dataset. 
Our approach improves over the prior art by a significant margin on all IoU thresholds except 0.7,
with a 10.3\% relative gain in mAP over the best baseline. A similar pattern can be observed 
from the ActivityNet-1.2 results summarized in Table~\ref{tab:anet2}. We can see that 
ASL improves over every baseline on all IoU thresholds except 0.5, with a 5.7\% relative gain in mAP.
These results indicate that the proposed action selection learning framework is highly effective 
for the weakly supervised temporal localization task. Figure~\ref{fig:ablate_class} further breaks down
THUMOS-14 performance by class. The top four classes with the largest relative improvement are highlighted 
in blue. The most improved classes have more than 100\% relative gain. We believe this is due to the
wide range of context settings present in the dataset for these classes, making class-agnostic action 
learning more effective for separating context. 

{\bf Ablation } To demonstrate the importance of actionness, we conduct ablation study on the THUMOS-14 
dataset shown at the bottom of Table~\ref{tab:thumos}. Here, $\asls$ uses class probability 
$h_{c,t} = s_{c,t}$ and $\asla$ uses actionness
probability $h_{c,t} = a_t$ for localization proposals during inference. The classification 
for $\asla$ is done at the video level by taking the class with the highest probability and
assigning it to every localization. Finally, ASL-BCE trains the actionness 
network $G$ with the binary cross entropy loss instead of the generalized 
$\mathcal{L}_{\text{ASL}}$ loss in Equation~\ref{eq:Lact}. We see that incorporating actionness
in the full ASL model relatively outperforms the classification-only $\asls$ approach by over 36\% in mAP.
Moreover, $\asla$ has very strong performance and is competitive with prior 
state-of-the-art results even though $a_t$ on its own has no explicit class information. This 
demonstrates that the actionness network $G$ is able to successfully capture the general
class-agnostic concept of action through our top-$k$ instance prediction task. Once captured, this property 
can be effectively used to identify regions within each video where the action occurs independently of the class.
We note here that attention models commonly used in prior work, have 
not been shown to be capable of localizing actions on their own. The full ASL model further improves 
performance of $\asla$ by 6\% indicating that classifier and actionness networks capture 
complementary information. Finally, using binary cross entropy instead of the
noise tolerant $\mathcal{L}_{\text{ASL}}$ loss hurts performance.
$\mathcal{L}_{\text{ASL}}$ becomes equivalent to binary cross entropy in the limit as 
$q \rightarrow 0$~\cite{zhang2018generalized}. In all our experiments we found 
that much higher values of $q$ such as $0.7$ produced better performance on both datasets, 
indicating that cross entropy is indeed not adequate here due to the high degree of noise in 
the target labels particularly at the beginning of training.

{\bf Actionness Learning } The main idea behind ASL is that actionness can be captured by predicting 
top-$k$ membership for each instance. In this section, we analyze this learning 
task in detail. Figure~\ref{fig:ablation} shows various properties of the $\tpos$ set throughout training.
In Figure~\ref{fig:ablation}\subref{fig:posk} we plot the fraction of instances in $\tpos$ that contain 
ground truth action from any target class over training epochs. For comparison,
we also plot this fraction for $\asls$ and $\asla$ where top-$k$ instances are chosen according
to class $s_{c,t}$ and actionness $a_t$ probabilities respectively.
We observe that without the actionness model, $\asls$ hovers around 63\% whereas for ASL it steadily
increases to over 72\%. Furthermore, $\asla$ reaches a much higher fraction of over 70\% compared
to $\asls$, capturing a significantly larger portion of instances with action. This again 
indicates that the actionness network is better at identifying action instances than the classifier.

Figure~\ref{fig:ablation}\subref{fig:aslbce} shows the validation accuracy of the actionness 
model $G$ in predicting which instances are in the $\tpos$ set. Despite the fact that
$\tpos$ is a moving target that can change with each iteration, the prediction accuracy 
remains stable and gradually improves throughout learning reaching over 84\%. The model is thus 
able to reach an equilibrium between the two networks and no divergence is observed.
Furthermore, Figure~\ref{fig:ablation}\subref{fig:iou} shows the intersection over union (IoU)
between the $\tpos$ sets from consecutive epochs during training. A higher IoU indicates a larger
overlap between consecutive $\tpos$ sets which in turn makes targets for $G$ more stable and 
easier to learn. We observe that the initial IoU starts around 0.5 and rapidly approaches 1 
as training progresses. Furthermore, the variance in IoU across training videos decreases 
throughout training so $\tpos$ sets stabilize after the first few epochs. 
These results indicate that the top-$k$ selection remains consistent for the majority of 
training epochs, and $G$ is able to successfully learn these targets; we observe this pattern 
in all our experiments.
\captionsetup[subfloat]{captionskip=-4pt}
\begin{figure*}[t]
    \vskip -0.4cm
    \begin{centering}
        \fbox{
            \subfloat[Hammer Throw]{\includegraphics[trim=0 0 0 0,clip,width=0.47\textwidth]{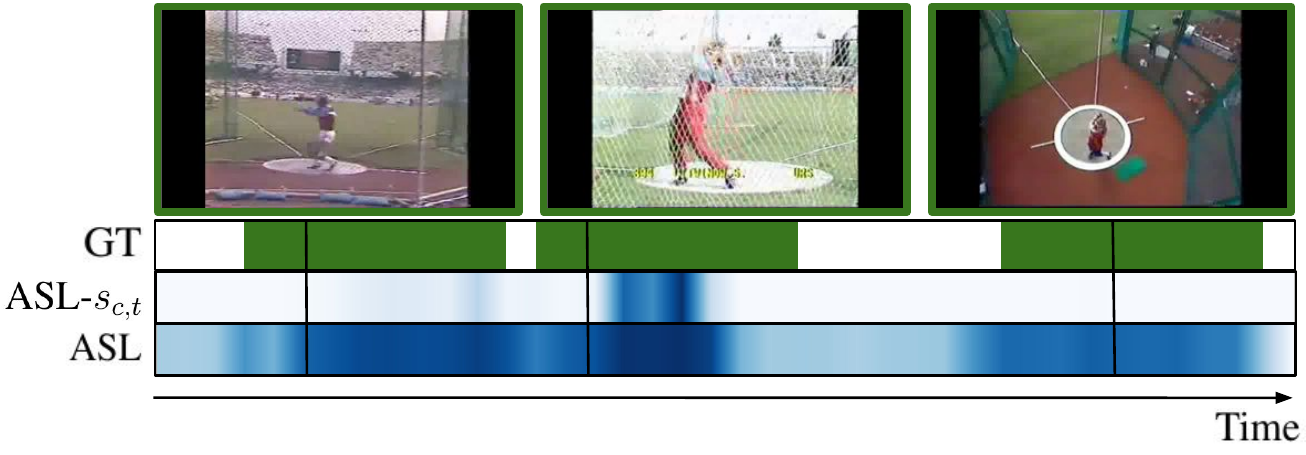}\label{fig:qual_case1}}
        }
        \fbox{
            \subfloat[Diving]{\includegraphics[trim=0 0 0 0,clip,width=0.47\textwidth]{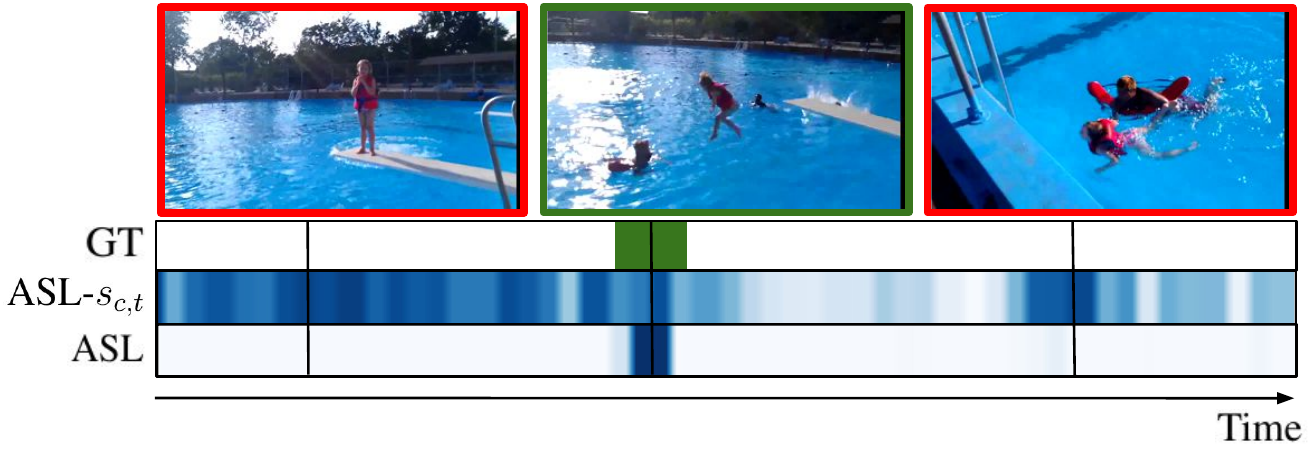}\label{fig:qual_case3}}
        }\\\vspace{-0.6cm}
        \fbox{
            \subfloat[Diving]{\includegraphics[trim=0 0 0 0,clip,width=0.47\textwidth]{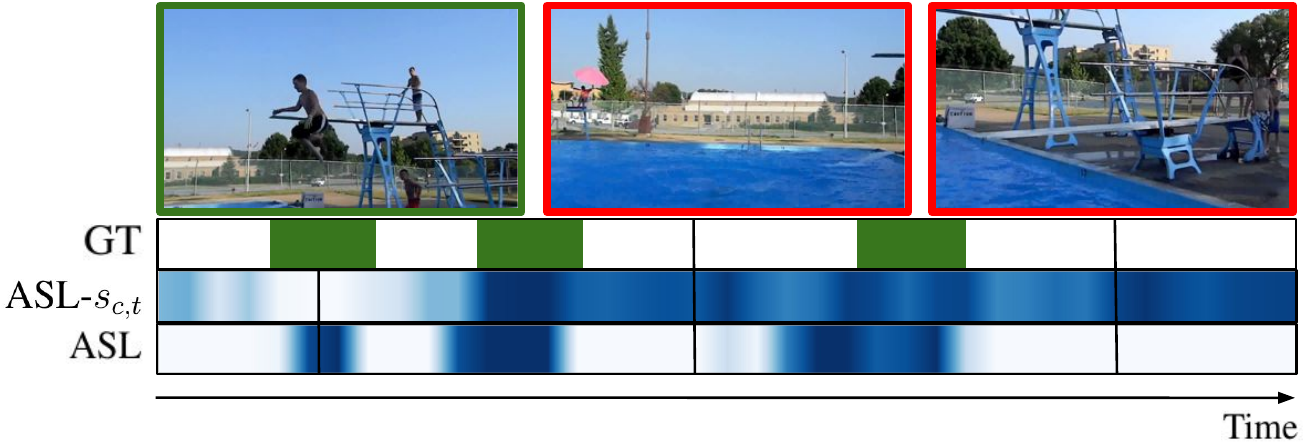}\label{fig:qual_case5}}
        }
        \fbox{
            \subfloat[Cricket Bowling]{\includegraphics[trim=0 0 0 0,clip,width=0.47\textwidth]{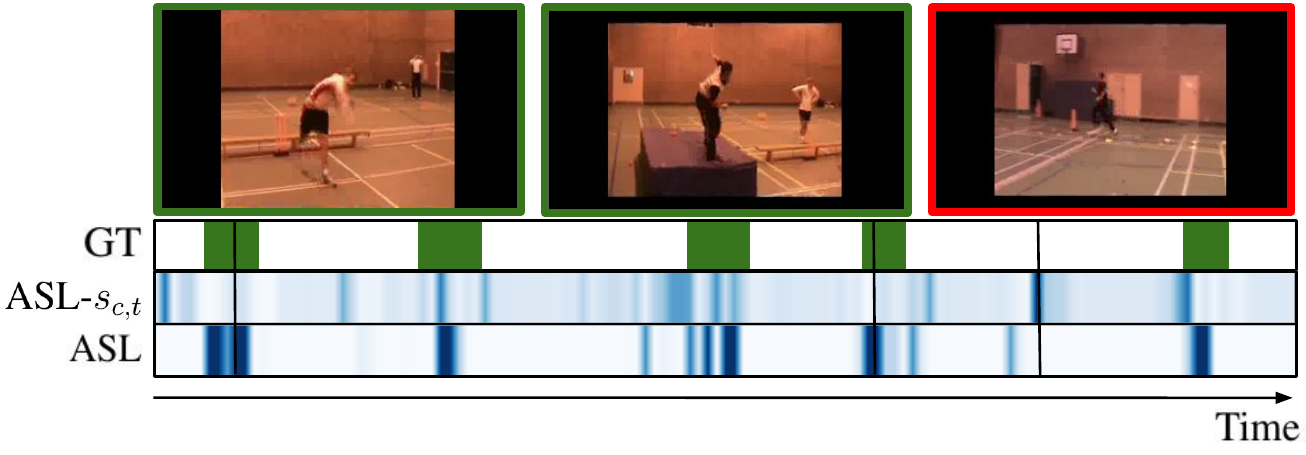}\label{fig:qual_case4}}
        }
        \vspace{-0.5cm}
        \caption{
        THUMOS-14 qualitative results comparing $\asls$ and ASL on four videos. Ground truth (GT) 
        localizations are indicated with green segments, and we show sample frames from
        action (green) and context (red) instances. Predictions are 
        shown in blue where darker color indicates higher confidence.
        }
    \label{fig:qualtitative_figure}
    \end{centering}
    \vspace{-0.6cm}
\end{figure*}

Figure~\ref{fig:error} breaks down the predictions made by ASL by error type. We use all test 
instances and show the total number of true positives (TP), false positives (FP), false negatives (FN)
and true negatives (TN). Note that we treat this as a binary problem and consider an instance as 
true positive if it contains an action and is selected for localization by ASL. In this setting, 
FP and FN represent context and actionness errors respectively, TP and TN are 
correctly predicted instances. We compare ASL with the classifier-only $\asls$ model and show
the relative increase/decrease for each category in blue. Figure~\ref{fig:error} shows that ASL 
improves each category predicting more instances correctly and making fewer mistakes. Specifically,
ASL reduces context and actionness errors by 31\% and 16\% respectively.

In Figure~\ref{fig:ablation} we showed that ASL can capture actionness because a significant proportion 
of instances in $\tpos$ sets contain actions, and these sets remain stable throughout training.
Moreover, even though not all instances in $\tpos$ contain actions, our noise tolerant loss 
is robust and can still perform well when a portion of labels is incorrect. Here, we further investigate 
the degree of noise that can be tolerated in this setting. Specifically, we evaluate
the ability of $G$ to capture actionness when it is trained on $\tpos$ with different proportions of 
action instances. Throughout training, we sample instances from $\tpos$ sets to lower the fraction
of instances that contain actions. This simulates a challenging learning environment where $G$ has 
to learn from increasingly noisier labels. Figure~\ref{fig:recall} shows these results on the 
THUMOS-14 dataset where we reduce the fraction of instances containing actions in $\tpos$ (Class Rate) from 70\% to 30\%. 
For comparison, the actual class rate on this dataset reaches around 72\% (see Figure~\ref{fig:ablation}\subref{fig:posk}). 
To more directly evaluate the impact of this setting we compute the instance-level Recall@100 using 
only predictions from $G$. Recall@100 is computed by measuring the fraction of instances that contain any action
amongst the top-100 instances predicted by $G$. We can see that Recall@100 for the noise-tolerant
ASL setting remains relatively stable between 50\% and 70\% but starts to drop significantly below
40\% class rate. These results suggest that the top-$k$ selection strategy can tolerate a high degree of noise 
with up to 50\% of incorrect labels. However, this also implies that the classifier needs to be sufficiently 
accurate in the top-$k$ selection for ASL to work.

Throughout the training, we simultaneously update both $F$ and $G$. We discussed that this results in 
moving targets for $G$ where instances in $\tpos$ change as the classifier is updated. Alternative 
training strategies are explored in Figure~\ref{tab:fgschedule}. We experiment with first training 
$F$ to convergence and then $G$ ($F$ then $G$), and alternating between training 
$F$ and $G$. We denote these alternating schedules by $F$:$n$ $G$:$m$ to indicate training 
$F$ for $n$ epochs followed by training $G$ for $m$ epochs and repeating. We observe that training 
$F$ then $G$ results in the lowest performance, since the model cannot adjust the classifier to work 
better with the actionness model. Alternating between $F$ and $G$ updates improves performance but still 
lags behind joint training. This corroborates our intuition that the classifier and actionness 
models complement each other in ASL and should be trained together.

\begin{figure}%
\centering
\vspace{-0.09cm}
\resizebox{0.95\linewidth}{0.7cm}{\includegraphics[clip, trim=0 12.5cm 0cm 0cm]{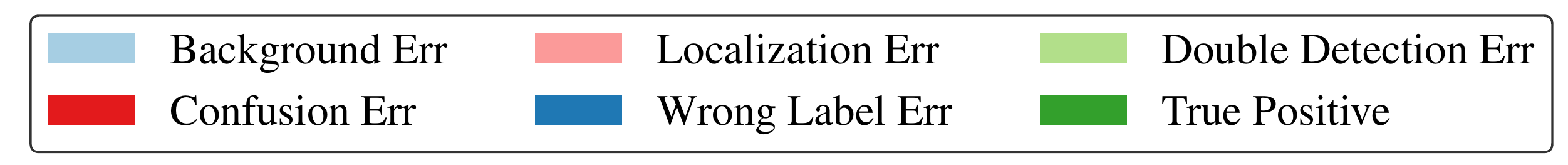}}\\\vspace{-0.04cm}
\hspace{-0.8cm}
    \setlength\tabcolsep{1pt} 
    \newcommand{\fnheight}{0.7\linewidth}
    \newcommand{\fnwidth}{0.67\linewidth}
    \newcolumntype{C}[1]{>{\centering\arraybackslash}m{#1}}
    \begin{tabular}{C{0.34\linewidth} | C{0.34\linewidth} | C{0.34\linewidth}}
    \resizebox{\linewidth}{!}{\includegraphics[clip, trim=0 0.2cm 11.5cm 2.7cm]{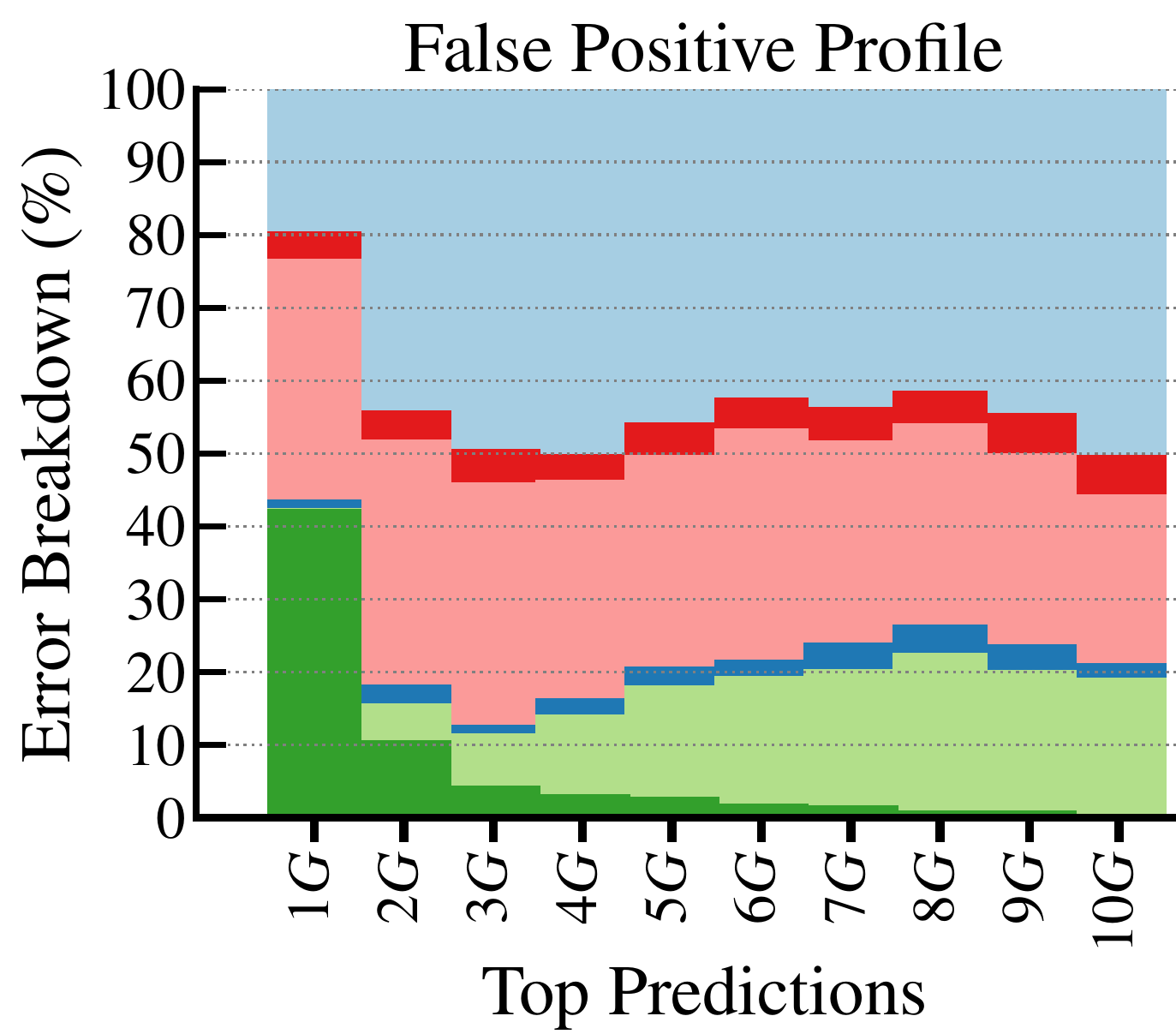}}
    &
    \resizebox{\linewidth}{!}{\includegraphics[clip, trim=0cm 0.2cm 11.5cm 2.7cm]{figs/false_positive_analysis_sct.pdf}}
    &
    \resizebox{\linewidth}{!}{\includegraphics[clip, trim=0cm 0.2cm 11.5cm 2.7cm]{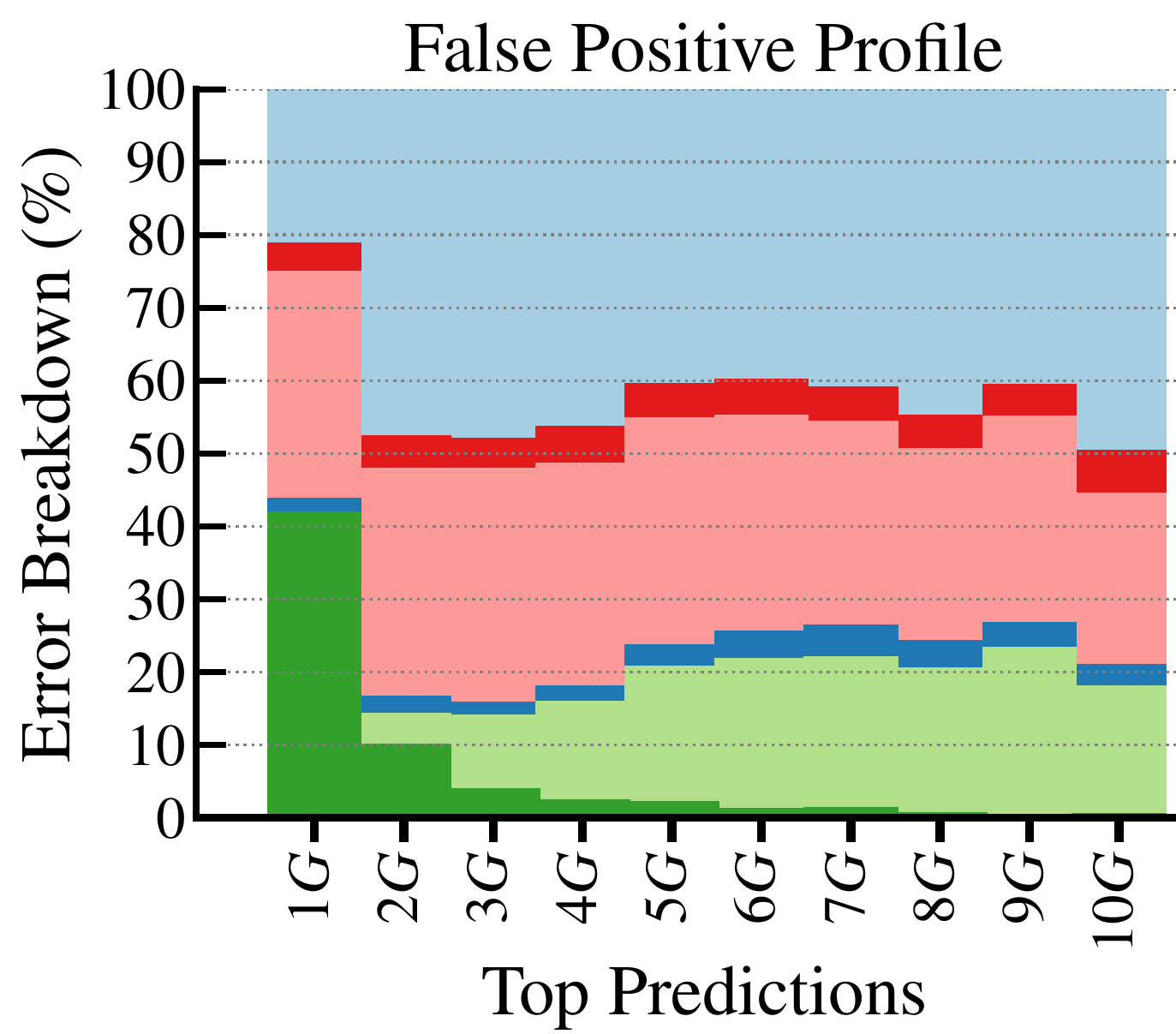}}\\\vspace{-0.15cm}
    \resizebox{!}{\fnheight}{\includegraphics[clip, trim=0cm 0.2cm 22.5cm 0cm]{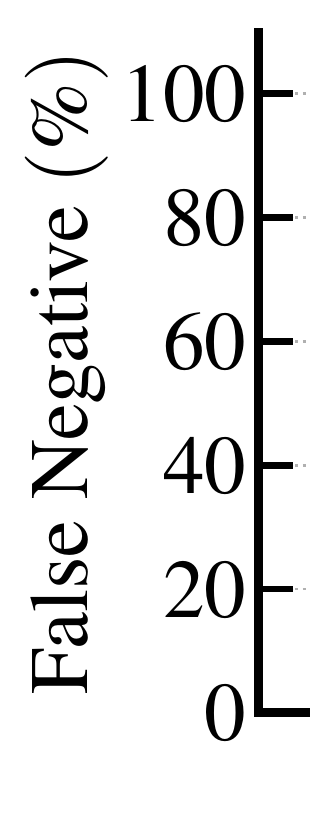}}
    \hspace{-0.5em}
    \resizebox{\fnwidth}{\fnheight}{\includegraphics[clip, trim=11cm 0.2cm 7.5cm 0cm]{figs/false_negative_analysis_asl.pdf}}
    &\vspace{-0.15cm}
    \resizebox{!}{\fnheight}{\includegraphics[clip, trim=0cm 0.2cm 22.5cm 0cm]{figs/false_negative_analysis_asl.pdf}}
    \hspace{-0.5em}
    \resizebox{\fnwidth}{\fnheight}{\includegraphics[clip, trim=11cm 0.2cm 7.50cm 0cm]{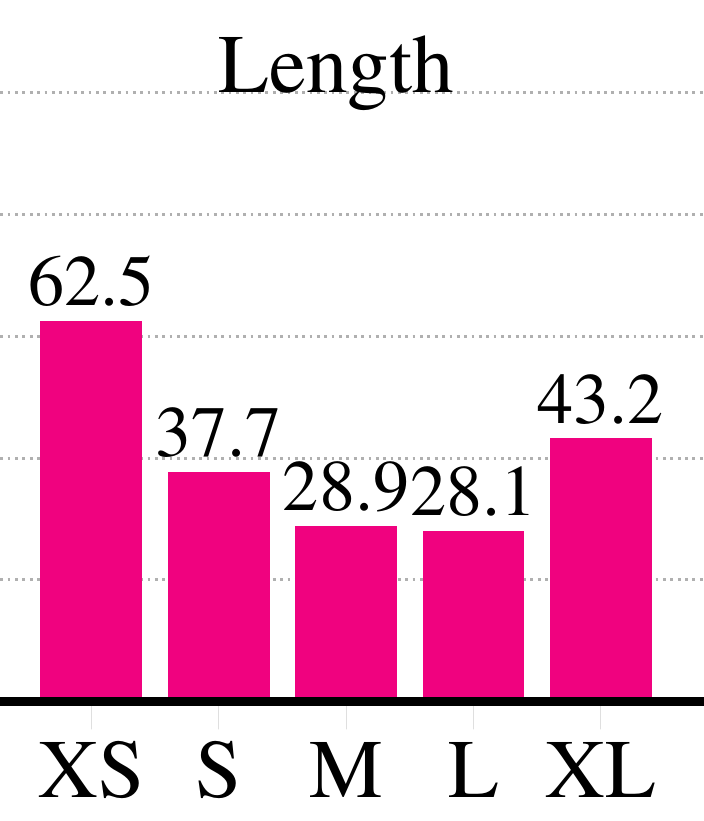}}
    &\vspace{-0.15cm}
    \resizebox{!}{\fnheight}{\includegraphics[clip, trim=0cm 0.2cm 22.5cm 0cm]{figs/false_negative_analysis_asl.pdf}}
    \hspace{-0.5em}
    \resizebox{\fnwidth}{\fnheight}{\includegraphics[clip, trim=11cm 0.2cm 7.5cm 0cm]{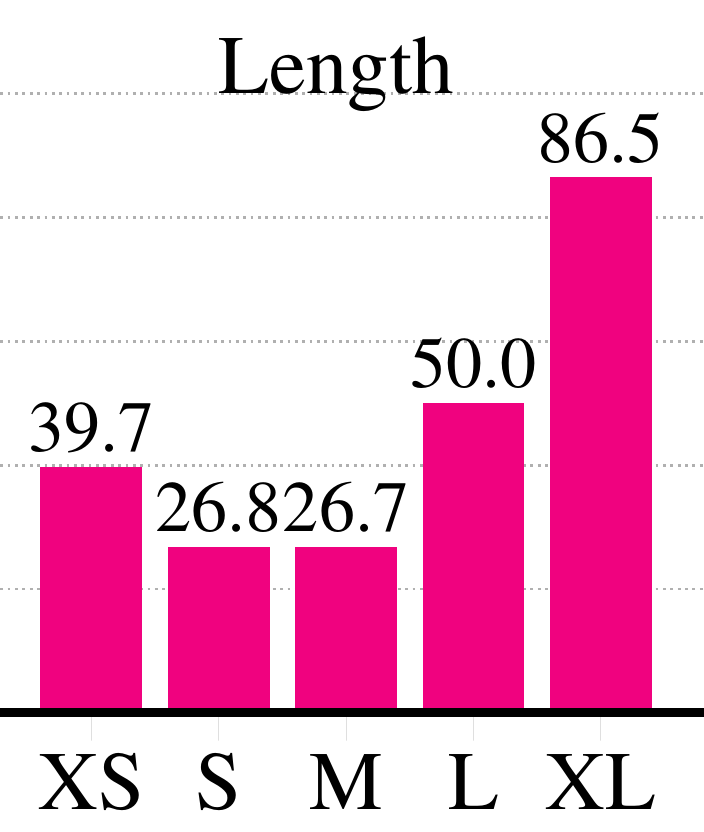}}\\
    [-0.2cm]
        \fbox{\subfloat[ASL\label{tab:detad_asl}]{\quad\quad\quad\quad\quad}}   &
        \fbox{\subfloat[$\asls$\label{tab:detad_asls}]{\quad\quad\quad\quad\quad}}  &
        \fbox{\subfloat[$\asla$\label{tab:detad_asla}]{\quad\quad\quad\quad\quad}} 
        \\
    \end{tabular}
\vspace{-0.4cm}
\caption{THUMOS-14 error analysis following DETAD \cite{alwassel2018diagnosing}. Top row breaks down false positives errors while bottom row shows false negatives by segment lengths.}
\vspace{-0.5cm}
\label{fig:detad}
\end{figure}

We further show DETAD \cite{alwassel2018diagnosing} analysis on THUMOS-14 dataset in Figure~\ref{fig:detad}. Here, the top row shows false positive analysis (context error) and 
false negative analysis (actionness error) is shown in the bottom row. The false positive profiles show that
$\asla$ significantly reduces the background (41.7 vs 47.7) and localization (28.3 vs 32.5) errors 
compared to $\asls$, further demonstrating that $G$ is able to learn actionness 
concepts that the classifier fails to capture. On the other hand, we observe higher double 
detection (14.8 vs 9.6) and wrong label (2.9 vs 0.9) errors in $\asla$ since it lacks 
class information. This corroborates our hypothesis that both models are needed to 
maximize localization accuracy. In the top predictions 1G, nearly all of the reduction in
background error translates to more true positives in both ASL and $\asla$ compared to $\asls$. 
On the false negatives, $\asla$ improves on shorter and more frequent action 
segments and complements $\asls$ which captures longer and more infrequent action segments better.

{\bf Qualitative Results} Figure~\ref{fig:qualtitative_figure} shows qualitative results 
for $\asls$ and ASL models. Ground truth segments are shown in green, and model predictions
are shown in blue with darker colors indicating higher confidence.
In figure~\ref{fig:qualtitative_figure}\subref{fig:qual_case1} for a video containing the ``Hammer throw" action, the $\asls$ localization alone is focusing on one very specific region of the video which likely contains the easiest 
instances (actionness error) to predict the video class. ASL spreads the localization predictions and correctly identifies all regions of actions. The opposite pattern is observed in Figures~\ref{fig:qualtitative_figure}\subref{fig:qual_case3} 
and~\ref{fig:qualtitative_figure}\subref{fig:qual_case5} which show the ``Diving" action class. 
$\asls$ activations are high on many context instances (context error) as they all 
contain highly informative scenes for the target ``Diving" class. ASL, however, focuses on the regions that contain the action. Lastly, Figure~\ref{fig:qualtitative_figure}\subref{fig:qual_case4} 
shows ``Cricket Bowling" action in an uncommon indoor setting. Here, 
$\asls$ has difficulty recognizing the action and most activations have low confidence. The actionness model is able to identify the action of throwing
as shown in the first two sample frames. Predictions of ASL are significantly more confident and identifies all of the action segments.

\section{Conclusion}

We propose the Action Selection Learning (ASL) approach for weakly supervised video 
localization. ASL incorporates a class-agnostic actionness network that learns a general
concept of action independent of the class. We train the actionness network with a novel 
prediction task by classifying which instances will be selected in the top-$k$ set by 
the classifier. Once trained, this network is highly effective on its own
and can accurately localize actions with minimal class information from the classifier.
Empirically, ASL demonstrates superior accuracy, outperforming leading recent benchmarks 
by a significant margin. Future work includes further investigation into actionness and 
its generalization to other related video domains.

{\small
\bibliographystyle{ieee_fullname}
\bibliography{main}
}
\end{document}